\documentclass[arxiv,12pt]{colt2026}

\newcommand{\cO}{\mathcal{O}}

\newtheorem{openproblem}{Open Problem}

\usepackage{mathtools, enumitem}

\title[Missing Inductive Bias for Algorithmic Compression]{Open Problem: Separating Geometric and Algorithmic Compression via Cayley-Table Completion}

\coltauthor{
\Name{Dongsung Huh} 
 }

\begin{document}

\maketitle

\begin{abstract}
Modern statistical learning theory and deep learning characterize generalization primarily in terms of continuous capacity control (e.g., norm-based regularization, margin maximization, low-rank bias). While highly successful in continuous domains, deep learning consistently fails to extrapolate exact algorithmic or discrete algebraic rules, reflecting a missing inductive bias toward algorithmic complexity minimization. We propose the Cayley-table completion as the canonical testbed for this missing bias, serving as the discrete algebraic counterpart to matrix completion. Just as matrix factorization combined with weight decay yields an implicit geometric bias toward low linear rank, recent results demonstrate that operator-valued tensor factorizations paired with a \textit{flatness prior} yield an implicit algorithmic bias toward exact discrete associativity. We pose the open problem of establishing formal exact recovery bounds for Cayley-table completion, and challenge the community to generalize continuous flatness priors to autonomously discover broader discrete algorithmic axioms without combinatorial search.
\end{abstract}


\section{Introduction: The Failure of Geometric Priors in Algebraic Domains}

The historical development of Statistical Learning Theory (SLT) reveals a deep theoretical divide between discrete logic and continuous geometry. In its classical formulation, learning was fundamentally framed as \textbf{Algorithmic Compression}. Classical frameworks such as PAC Learning for boolean concepts \citep{valiant1984theory}, Occam's Razor bounds \citep{blumer1987occam}, and Minimum Description Length (MDL) \citep{rissanen1978modeling} evaluated generalization based on the recovery of exact, minimal logical rules (e.g., boolean circuits, finite automata).

Because optimizing discrete algorithmic complexity is generally NP-hard, modern SLT largely abandoned it in favor of continuous capacity control mechanisms—an approach we term \textbf{Geometric Compression}. Geometric compression assumes that parsimonious underlying concepts lie on smooth manifolds or low-dimensional linear subspaces. This shift was theoretically justified by landmark results in fields like compressed sensing and matrix completion, which rigorously proved that continuous geometric penalties (like $\ell_1$ or nuclear norms) serve as perfect convex relaxations for discrete NP-hard properties (like $\ell_0$ sparsity or discrete matrix rank) \citep{fazelRankMinimizationHeuristic2001, candesExactMatrixCompletion2009}. Consequently, modern SLT assumes that implicit geometric biases toward smoothness, margin maximization, and low linear rank are sufficient for robust generalization \citep{srebro2004maximum, gunasekar2017implicit}.

This geometric paradigm fundamentally underpins generalization in modern deep learning.
Deep neural networks excel at discovering low-rank embeddings and smooth interpolations within low-dimensional latent spaces \citep{bengio2012representation, ansuini2019intrinsic, aghajanyan2021intrinsic, huh2021low, morwani2023simplicity}, which proves remarkably effective for representing natural, continuous data. However, this geometric surrogate relationship categorically fails in discrete algebraic domains. This theoretical misalignment manifests empirically across modern architectures. Standard neural networks categorically struggle to achieve true length generalization on algorithmic tasks \citep{schwarzschild2021can}, and require massive, brittle memorization rather than rule abstraction to simulate deterministic finite automata \citep{deletang2023neural}.

In this open problem, we propose \textbf{Cayley-table completion} as the canonical testbed for learning discrete algebraic structures. In this setting, geometric compression is fundamentally at odds with algorithmic compression, exposing a crucial missing inductive bias in the current paradigm. 

\paragraph{Motivating Example: Deep Learning's Inability to Discover Exact Symmetries.}
A striking manifestation of this missing inductive bias occurs in the context of Geometric Deep Learning (GDL) \citep{bronstein2021geometric}. When standard neural networks are trained on symmetry-augmented datasets, they learn only \textit{approximate} symmetries, consistently failing to natively discover the exact weight-sharing patterns of the underlying symmetry groups. This exposes a fundamental limitation: deep learning lacks the inductive bias to algorithmically compress these continuous approximations into rigid, shared weight structures. Instead, to guarantee out-of-distribution (OOD) generalization, GDL imposes the exact weight-sharing patterns of an \textit{a priori} known symmetry group via hard-coded equivariant architectures \citep{cohen2016group}.

\section{Canonical Tasks}


While the domain of algorithmic compression is vast—spanning formal languages to general automata—establishing a rigorous foundation requires a minimal atomic testbed. 
Just as Matrix Completion serves as the foundational testbed for the discovery of continuous low-dimensional manifolds, Cayley-Table Completion---the recovery of a discrete binary operation from partial observations \citep{huh2025discovering, powerGrokkingGeneralizationOverfitting2022}---provides the canonical formalization for discovering discrete algebraic rules.

This frames a fundamental dichotomy in structural recovery:
\begin{itemize}
\item \textbf{Matrix Completion:} The underlying structure has \textit{low linear rank}. The continuous solution is achieved via Matrix Factorization paired with an $\ell_2$ penalty. Factoring a matrix $M = AB$ and applying weight decay ($\ell_2$ regularization) yields a variational characterization of the nuclear norm of $M$:
$$\min_{A,B \text{ s.t. } AB = M} \| A \|^2_F + \| B \|^2_F = 2 \sum_i \sigma_i(M),$$
which serves as a differentiable (but \emph{inexact}) surrogate for linear rank \citep{gunasekar2018implicit,shang2020unified,dai2021representation}.
\item \textbf{Cayley-Table Completion:} Cayley tables of groups (and quasigroups) are Latin squares, whose rows are permutations of one another, thus inherently precluding linear redundancy. Instead, the true parsimony of algebraic groups stems from the associativity axiom, which tightly constrains the global structure to yield remarkably low algebraic complexity, effectively minimizing its structural description length. 
The fundamental challenge, therefore, is finding a continuous, differentiable measure that can penalize associativity violations without resorting to intractable combinatorial search.
\end{itemize}

\section{Existence Proof: Differentiable Measures of Algebraic Complexity}

To formalize this mathematically, let $Q$ be a finite set of $n$ elements equipped with a binary operation $\circ: Q \times Q \to Q$. The operation's Cayley table is captured by the structure tensor $\delta \in \{0,1\}^{n \times n \times n}$, where $\delta_{abc} \coloneqq \mathbb{I}_{\{a \circ b = c\}}$ for $a,b,c \in Q$.

Recently, \citet{huh2025discovering} 
demonstrated that Cayley-table completion can be exactly solved via an operator-valued tensor factorization
$$~~~~~~~~~~~~ T_{abc}(\Theta) \coloneqq \frac{1}{n} \operatorname{Tr}(A_a B_b C_c),  ~~~~~~~~~~~\Theta = (A,B,C)$$
where $A_a, B_b, C_c$ are the $n \times n$ matrix slices (linear operators) indexed by $a,b,c \in Q$.
This factorization is optimized under a \textbf{flatness prior} that penalizes the Hessian trace of the empirical reconstruction loss: 
$$\mathcal{H}(\Theta) \coloneqq \operatorname{Tr} \left( \nabla^2 \sum_{(a,b,c)\in\Omega} (T_{abc}(\Theta) - \delta_{abc})^2 \right),$$
%
\citet{huh2025hypercube} provided a rigorous landscape analysis of this objective, proving that in the fully-observed limit, the geometric infimum $\mathcal{H}_{\inf}(\delta) \coloneqq \inf_{T(\Theta)=\delta} \mathcal{H}(\Theta)$ implicitly defines an exact, differentiable measure of algebraic complexity.%
%

Crucially, the objective $\mathcal{H}(\Theta)$ decomposes into two components: a geometric alignment term (collinearity) and an \emph{inverse} $\ell_2$ penalty. These variational terms strictly induce core discrete properties. Collinearity enforces associativity, while the inverse $\ell_2$ term acts as an exact inverse rank penalty within the collinear manifold. This mechanism actively drives the parameters toward full-rank unitarity, diametrically opposing the ubiquitous low-rank bias of standard geometric compression.

Consequently, this geometry establishes an absolute lower bound $\mathcal{H}_{\inf}(\delta) \ge 3 |\delta|$ (where $|\delta| \coloneqq \sum_{abc} \delta_{abc} = n^2$ denotes the table size). This bound is attained \emph{if and only if} the target operation is isotopic to a group, uniquely characterizing the global minimizer as the regular representation of the underlying group (up to unitary gauge). This establishes a formal proof-of-concept that continuous gradient-based optimization can natively discover discrete algebraic axioms, bypassing combinatorial search entirely.

\section{Open Problems in Algorithmic SLT}

These partial results suggest that exact algorithmic complexity can be natively optimized via continuous geometric penalties. Empirically, correcting this missing inductive bias yields drastic structural benefits: \citet{huh2025discovering} demonstrated that this flatness-regularized tensor factorization does not merely approximate the target operations, but rapidly collapses into exact recovery, achieving state-of-the-art sample efficiency that scales tightly with the true algorithmic complexity of the underlying group. 

However, the foundational question remains: what is the exact theoretical boundary between generalization via geometric compression and algorithmic compression? To formally map the territories that only algorithmic compression can achieve, we pose the following open problems.

\begin{openproblem}[Formal Separation of Geometric and Algorithmic Compression]
The core theoretical challenge is to establish a strict separation theorem between geometric capacity control (e.g., norm-based regularization, Rademacher complexity, margin bounds) and algorithmic compression. Can we formally characterize the hypothesis classes—such as discrete algebraic rules or formal languages—where all purely geometric generalization bounds are provably vacuous (requiring exponential sample complexity), yet differentiable algorithmic complexity measures yield tight, polynomial guarantees? Mapping the exact territories where geometric surrogates fundamentally fail, and where only algorithmic compression succeeds, remains a critical missing foundation in SLT.
\end{openproblem}

\begin{openproblem}[Sample Complexity of the Canonical Testbed]
Cayley-table completion serves as the immediate, concrete testbed for establishing this separation. Because finite groups are linearly full-rank ($r=n$), applying standard matrix completion theory yields vacuous bounds exceeding $\cO(n^2)$. Can we formally establish an exact recovery guarantee showing that the global minimizer of the flatness-regularized empirical loss perfectly completes the group table with high probability given only $\cO(n \log n)$ uniformly sampled entries? Solving this would provide the first rigorous mathematical proof of the separation demanded in Open Problem 1.
\end{openproblem}

\section{Related Paradigms: Limitations of Current Neurosymbolic Approaches}

Recent neurosymbolic approaches attempt to bridge the gap between continuous optimization and discrete logic, but they generally fall into one of two methodological bottlenecks.

The first approach relaxes exact logic into continuous probabilities or fuzzy logic (e.g., Logic Tensor Networks \citep{serafini2016logic}, DeepProbLog \citep{manhaeve2018deepproblog}). While fully differentiable, these methods yield approximate probabilistic inferences rather than recovering rigid algebraic structures. The second approach uses deep learning merely as a heuristic guide for discrete combinatorial search (e.g., standard neural theorem provers, AlphaGeometry \citep{trieu2024alphageometry}). While capable of outputting exact structures, these methods rely heavily on external symbolic engines to handle the logical constraints, leaving the fundamental NP-hard barrier of native algorithmic discovery intact.

\section{Significance to Learning Theory}

For decades, learning theory has treated true algorithmic compression as computationally intractable, retreating to geometric capacity control as a tractable surrogate. Formalizing and solving the Cayley-table completion problem establishes a rigorous mathematical bridge between the classical logic-driven and modern geometry-driven eras of SLT. It provides a formal proof-of-concept that carefully designed, purely continuous priors can compel gradient descent to optimize exact combinatorial properties natively. Addressing these open problems is a vital step toward endowing neural networks with the inductive biases necessary to autonomously discover rigid mathematical and algorithmic laws.


\bibliography{COLT_references}

\end{document}